\documentclass{article} 
\usepackage{iclr2016_workshop,times}
\usepackage{hyperref}
\usepackage{url}
\usepackage[utf8]{inputenc}
\usepackage{graphicx,subcaption}
\usepackage{amsmath}
\usepackage{cases}
\usepackage{bbm}

\title{Learning to decompose for object detection and instance segmentation}

\author{Eunbyung Park \& Alexander C. Berg \\
Department of Computer Science\\
University of North Carolina at Chapel Hill\\
\texttt{\{eunbyung,aberg\}@cs.unc.edu} \\
}

\begin{document}

\maketitle

\begin{abstract}
Although deep convolutional neural networks(CNNs) have achieved remarkable results on object detection and segmentation, pre- and post-processing steps such as region proposals and non-maximum suppression(NMS), have been required. These steps result in high computational complexity and sensitivity to hyperparameters, e.g. thresholds for NMS. In this work, we propose a novel end-to-end trainable deep neural network architecture, which consists of convolutional and recurrent layers, that generates the correct number of object instances and their bounding boxes (or segmentation masks) given an image, using only a single network evaluation without any pre- or post-processing steps. We have tested on detecting digits in multi-digit images synthesized using MNIST, automatically segmenting digits in these images, and detecting cars in the  KITTI benchmark dataset.  The proposed approach outperforms a strong CNN baseline on the synthesized digits datasets and shows promising results on KITTI car detection.
\end{abstract}

\section{Introduction}

State-of-the-art object detection methods are based on a combination of region proposals (possibly using a deep network) and deep convolutional neural networks(CNNs) for classification. The currently typical recipe is to propose many object candidate boxes based on an objectness measure and classify each proposed box. This first became standard practice with work from \citet{vandeSandeICCV2011}, and is now part of the best performing systems, using deep CNNs as classifiers\citep{girshick14CVPR,girshickICCV15fastrcnn,ren2015faster}, sometimes using deep networks for proposing candidate reagions as well. Post-processing, such as non-maximal suppression, is performed to prune out many false positives and is known to be one of the critical factors that affect performance\citep{Wan_2015_CVPR}.  The tension between more object proposals in order to cover any possible object and the difficulty of later pruning out high-scoring but incorrect bounding boxes---and the computational cost of evaluating all the candidate boxes---is one of the challenges in current detector design.  Some work, like \citet{ren2015faster} tried to solve this problem by using a deep CNN to generate a relatively small number of high-quality candidates, but still require several hundreds of box evaluations. Another tack, e.g., \citet{RedmonDGF15}, removes the region proposal step and directly predicts multiple objects and their locations with a single CNN evaluation. However that method is restricted to a predefined number of outputs (e.g. 49 in a 7x7 grid) even when there are fewer or more objects present, and its performance falls short compared to previous region-based detectors.  Our aim is to produce a network that generates the correct number of object instances and their bounding boxes (or segmentation masks) given an image, using  only a single network evaluation without any pre- and post-processing steps, such as region proposals and non-maximal suppression. Building these together into a single framework enables much more straightforward end-to-end training of multiple-class multiple-object detectors, potentially allowing easier and wider application, as well as better performance.

Some recent results address sub-components of this problem and indicate that a solution may be possible.  Predating the region proposal and deep-network-based classifier evaluation approaches, \citet{alexnet} showed that their network applied to a whole image had enough information to predict a single object bounding box location per image.  Looking into spatial information contained in activation maps at intermediate layers in deep networks, work from \citet{Simonyan14,Zhou15} and others have shown the ability to localize objects in images.  However, these works either do not handle multiple-object detection (instead only producing a single detection per image), or rely on iteratively evaluating a further classifier over possible bounding boxes (chosen based on the activation map).  Other approaches try to directly count objects without explicitly addressing recognition or localization \citet{Zhang_2015_CVPR}.  Toward general computational savings, \citet{ren2015faster}, showed deep CNN features can be shared for both object proposals and classification. The approach in this paper builds on several of these ideas to move from a single detection or a fixed number of detections per image to a full multi-class and variable number of instances detector that can also produce instance segmentations.



\begin{figure}[t]
\centering
\includegraphics[width=\linewidth]{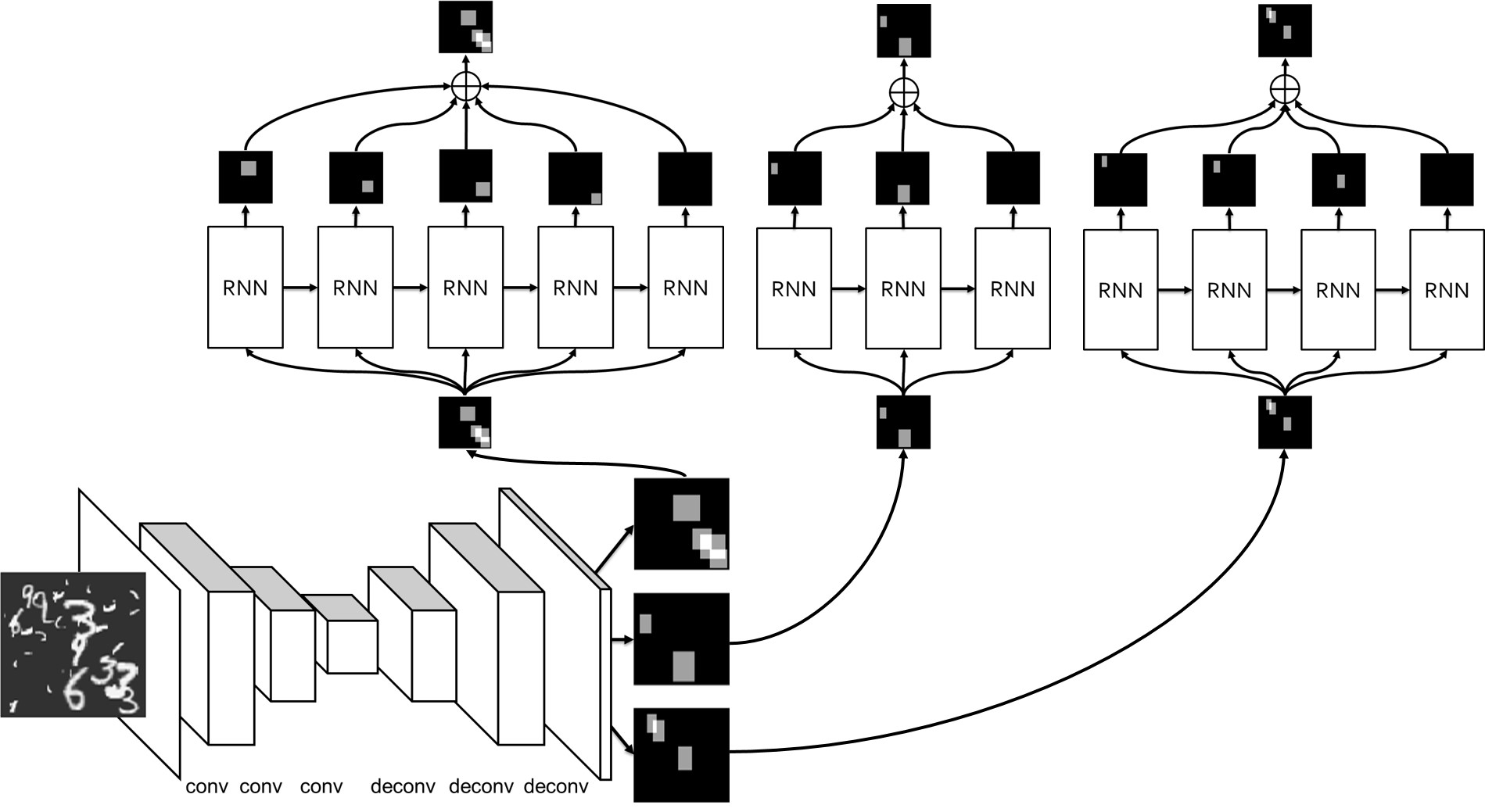}
\caption{Overall architecture of \textit{decompNet}. The lower network processes an image into an activation map for each category(section \ref{sec:category_decomposition}). The upper part of the network decomposes each cagtegory-specific activation map into multiple instances(section \ref{sec:instance_decomposition}). We synthesized a detection dataset from MNIST using digits 3, 6, and 9 (section \ref{sec:mnist_dataset}).}
\label{fig:decompnet}
\end{figure}

We present a novel end-to-end trainable deep neural network architecture consisting of both convolutional and recurrent layers for multi-class multi-object detection and instance segmentation with a single network evaluation. Object detection is a function of an image that produces multiple bounding box locations and category labels.  The number of objects detected and their categories can vary between images. Recurrent layers play a critical role in making structure of network produce variable length of outputs according to the inputs. The structure of our proposed network is shown in Figure~\ref{fig:decompnet}.  An input image is processed to produce multi-scale activation maps for each of the target object categories.  Each of these activation maps is fed into a recurrent network that produces a variable number of object detection maps that mask out the location of individual instances.  In order to train this network, we proposed a new mask-based loss function that considers precise individual instance location and the total number of object instances simultaneously, allowing the proposed network to produce a the number of object instances appropriate for each image.

\section{DecompNet}

In this section, we describe the proposed network architecture, \textit{decompNet}. Fundamentally, it differs from previous work in two aspects. First of all, \textit{decompNet} produces object category response maps and decomposes them into multiple individual instance maps. This is the opposite of previous methods. Most previous work on object detection first finds object candidates, and then determines the object category\citep{girshick14CVPR,Erhan_2014_CVPR}. This approach can require a great deal of computation since it has to go through classification for every object candidate.  Our decomposition is relatively light-weight compared to the classification process.

Secondly, the decomposition stages of \textit{decompNet} look at the entire response map and each stage of the recurrent network can return a single instance. This is also very different from existing methods that divide input images into spatial grids and produce one or multiple predictions per cell by only looking at the local cells in the grid. Region proposal network in \citet{ren2015faster} produces k region proposals per each cell. \citet{RedmonDGF15} produce one object per one location of a 7x7 output grid.
\citet{Stewart15} produce multiple faces of people per cell of the last convolutional feature maps. We believe this is mainly because there is an architectural limitation of feed forward CNNs and dividing the image into spatial grid with multiple bins is a simple solution to design feed forward CNNs to produce multiple outputs. These works might have difficulty recognizing larger objects since it is known that the empirical size of receptive fields are not large enough\citep{Zhou15}. They also might have difficulty when there are objects in same cell location.

\subsection{Category decomposition}
\label{sec:category_decomposition}
The goal of the first part of the network is to produce response maps of each object category. We keep high resolution response maps because they will try to give precise location and size of objects to the second part of the network. 

The proposed network, depicted in figure \ref{fig:decompnet}, is an all convolutional network inspired by \citet{Springenberg15,Lin13,noh2015learning}. We didn’t use any fully connected layers since they only increased the number of parameters without notable performance improvement. We used strided convolution instead of max-pooling for downsampling and used deconvolution layers for recovering the resolution back to half or original size of the input image. At the final layers, we put a 1x1 convolution with the same number of filters as the number of categories. Thus, the outputs of the final convolutional layer are response maps for each category. We adopted a mask based loss function for penalizing activations at the location where there is no object corresponding to the category \citep{Szegedy13}.

\begin{equation}
\sum_{i=1}^{N}\sum_{c=1}^{|C|}{D_{\lambda}(f^{c,i},\sum_{t=1}^{T(c,i)}m_{t}^{c,i})}
\end{equation}

The network is trained over N training examples with modified L2 distance between response map $f^{c,i}$ and sum of instance ground truth masks $\sum_{t=1}^{T(c,i)}m_{t}^{c,i}$, where $m_{t}^{c,i} \in \big\{ 0,1 \big\}^{M}$ is the t\textsuperscript{th} instance ground truth mask for category c. $T(c,i)$ returns the number of instances of in the category $c$ in the i\textsuperscript{th} image. The distance function is defined
\begin{equation}
D_{p}(x,y)=||\sqrt{Diag(y)+pI}(x-y)||
\end{equation}

As mentioned in \citet{Szegedy13}, $\sqrt{Diag(y)+pI}$ term played a key role in avoiding trivial solutions producing all zero outputs. The most important part was the fact that we encoded overlapped regions of instances in the loss function. Every individual instance mask is a binary mask. We sum all of the individual masks, which results in higher values at the overlapped regions so that we can make the second part of network aware of the presence of multiple objects at the location. This was very critical to the performance.

\subsection{Instance decomposition}
\label{sec:instance_decomposition}
Once we have category response maps the second part of the network, which is shown in the upper part of the figure~\ref{fig:decompnet}, takes response maps and produces response maps of the same size that only contain one instance at a time. The category response map consists of several blobs of responses and each blob represents one instance of the category. Thus this network is responsible for splitting several response blobs into one individual blob. The job of this network is relatively simple, as classification and localization, have been performed by the former network. Thus, the computational cost of the recurrent network is not too expensive despite havin  several stages to allow detecting multiple instances.

\begin{figure}[t]
\centering
\includegraphics[width=\linewidth]{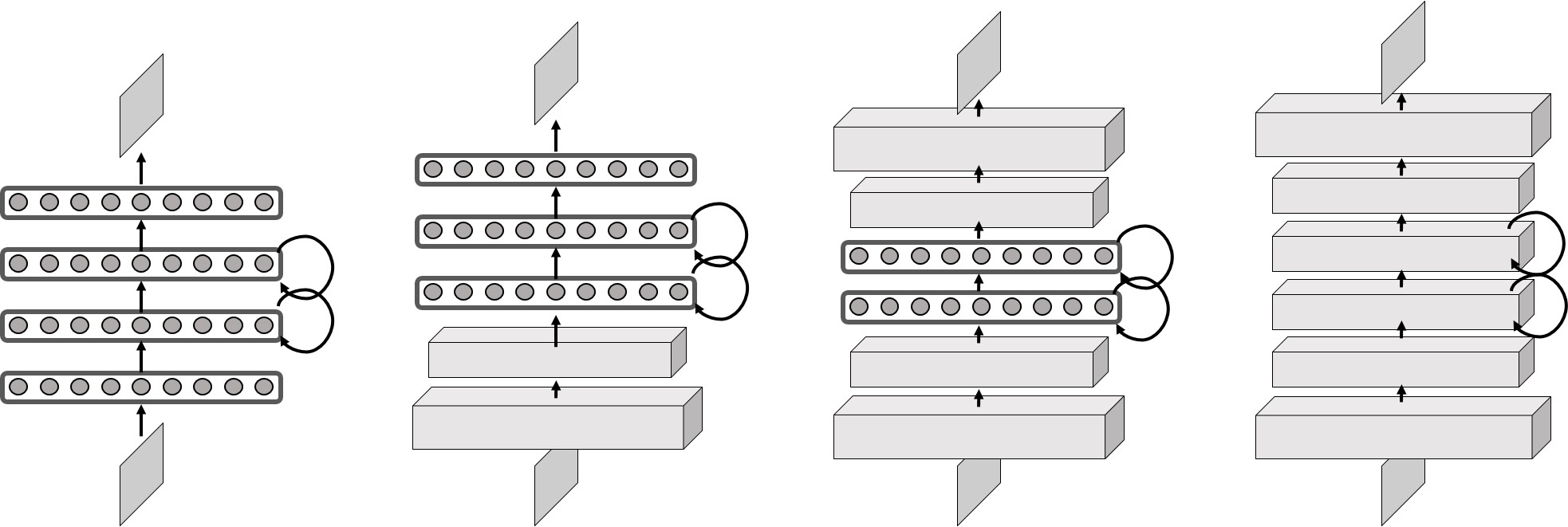}
\caption{Various RNN alternatives. The first one is only based on fully connected layers. In the second network, we used convolution for first two layers to reduce the input resolution and other layers are constructed by fully connected layers. In the third network, we used convolutional layers except for the bottleneck fully connected recurrent layers. We also tried all convolutional recurrent network dipicted in the last column.}
\label{fig:rnns}
\end{figure}

We used a recurrent neural network in order to realize two important aspects of this job. The network should be able to produce variable length outputs and memorize previous states so that it can generate individual response blobs that it has not produced so far. Figure \ref{fig:rnns} shows several alternatives of network architectures. The first and second require many parameters and do not give us any better results. The third and fourth give us similar performance and we used the third one for all of our experiments. The intuition behind this network is the following. We can think of each blob as an object in the image. It will recognize individual blobs from the first one or two convolutional layers and generate instance response map with the last one or two convolutional layers. With bottleneck recurrent layers, the network will memorize the blobs that were already generated and give new blobs to higher convolutional layers.

\begin{equation} \label{eq:3}
\sum_{i=1}^{N} \sum_{c=1}^{|C|} \bigg[ \sum_{t=1}^{T(c,i)} D_{\lambda}(g_{t}^{c,i}, m_{idx(t)}^{c,i}) + \eta D_{\gamma}( \sum_{t=1}^{T(c,i)}g_{t}^{c,i}, \sum_{t=1}^{T(c,i)}m_{t}^{c,i}) \bigg]
\end{equation}

The loss function consists of two distance terms, $D_\lambda$ and $D_\gamma$. $\lambda$, $\gamma$, and $\eta$ are hyperparameters. The $idx()$ function returns the corresponding id of ground truth masks. We find the best bipartite matching between outputs of network and ground truth masks based on L2 distance. The first distance term is for comparing individual instances with ground truths. This term will penalize response maps that have more than one instance blob. The second distance term works as a stop condition of the recurrent network. This term is the distance between sum of instance maps that the network has generated and sum of ground truth masks. So, it will penalize the network if the network produced more instances than the ground truth. It turns out that this term was very important. The network produced a lot of instance maps and didn't know when to stop without this term, which resulted in very low precision for object detection tasks.

\subsection{Bounding box regression}
We performed bounding box regression given each instance response map. It predicts a bounding box as tuple $b_{\text{est}}$ = ($x$,$y$,$w$,$h$) where $x$ and $y$ are center coordinates of the box, and $w$ and $h$ are width and height respectively. $x,y,w,h$ are each scaled from 0 to 1. We could have written a simple program for drawing bounding boxes given each instance response map, but there is sometimes noise in response maps. So, We applied simple regression based method with dropout that gave us robust bounding box prediction. We used a simple feed forward neural network that was constructed by 1 hidden layer with 4K hidden units. We strictly followed the smooth L1 loss function in \citet{girshickICCV15fastrcnn}.
\begin{equation}
\text{smooth}_{L1}(x,y) = \begin{cases} 
0.5(x-y)^2 & \text{if $|x-y|<1$} \\
|x-y|-0.5 & \text{otherwise,}
\end{cases}
\end{equation}
We computed a score for each detected object based on agreement between estimated bounding box and intensity values in the instance map produced by our network. Given an instance map $g$ and corresponding bounding box estimation $b_{\text{est}}$ we defined score function as following,
\begin{equation}
\text{score}(g,b_{\text{est}}) = g^\top M(b_{\text{est}}) / A(b_{\text{est}})
\end{equation}
, where $M(b_{\text{est}})$ and $A(b_{\text{est}})$ returns the vectors of binary mask and area of bounding box estimation respectively.

\subsection{Instance segmentation}

Recently, object instance segmentation has gained much attention since people have collected large-scale instance segmentation datasets\citep{Liang15,BharathECCV2014,mscoco}. Unlike semantic segmentation, individual instances of an object category have to be segmented out separately. One good thing about \textit{decompNet} and an our proposed mask-based loss function is that they can directly apply to instance segmentation without any modification. Our network produces activation values close to 1 at confident place and close to 0 at low confident place. Thus, we defined a score as average activation over a instance map
\begin{equation}
\text{score}(g) = \sum_{i} g_i \mathbbm{1}{\left[g_i > \delta\right]}  \Big/  \sum_{i} \mathbbm{1}{\left[g_i > \delta\right]}
\end{equation}
, where $g$ is a instance map that the network produces at each time step and $\delta$ is threshold value for ignoring noise activation values.

\section{Experiments}

\subsection{Synthesized dataset}
\label{sec:mnist_dataset}

\begin{figure}[t]
\centering
\includegraphics[width=\linewidth]{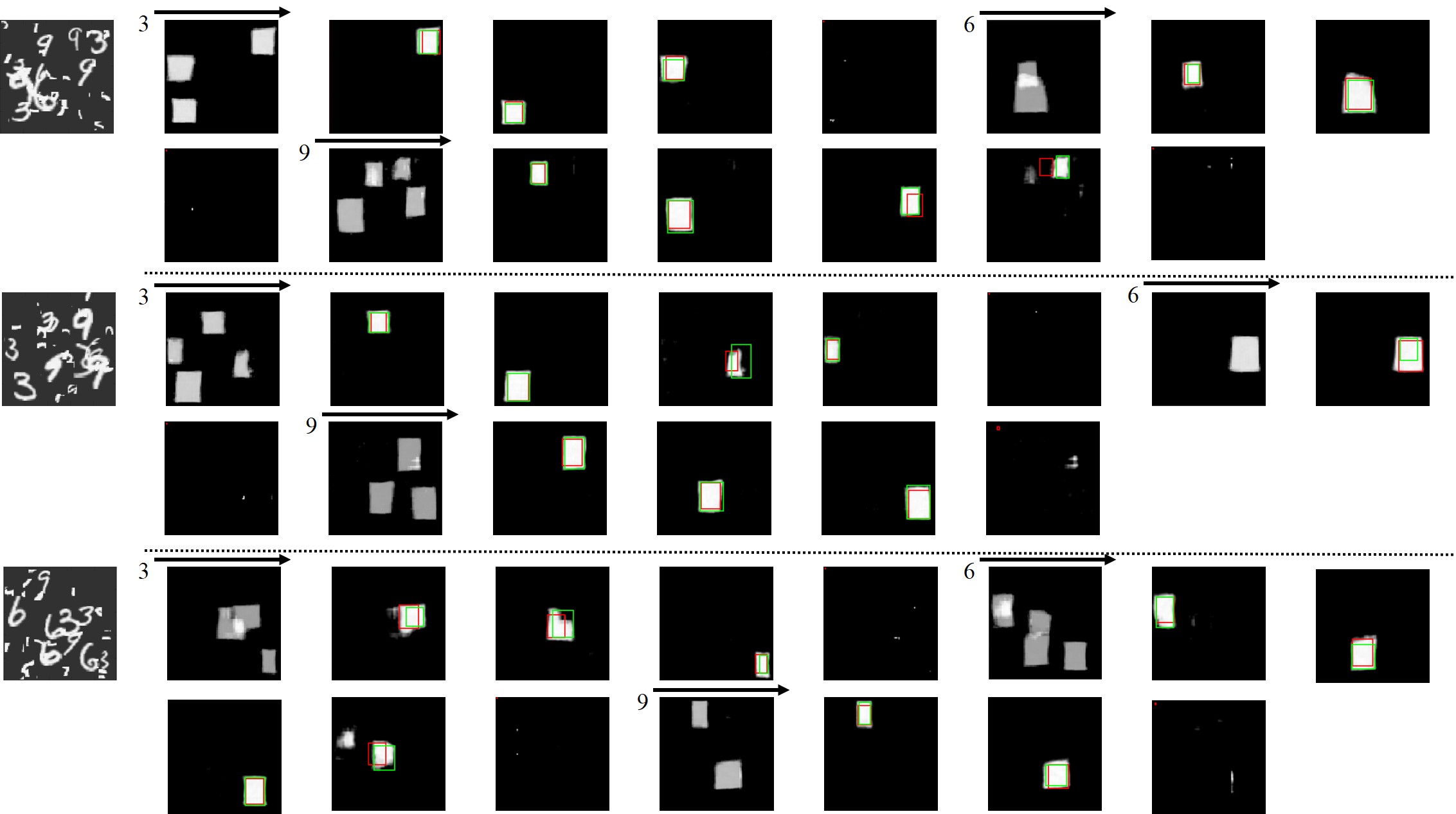}
\caption{Examples of object detection results: The leftmost column is input images. The instance maps are presented in the order of digit 3, 6, and 9. The first response map is the sum of all instance maps for each category. Therefore, it contains all instances that our network produced. And every instance map is arranged in the same order as produced by the network. The last instance map for each category is usually a black image and it works as a stop sign. Red boxes are the outputs of bounding box regression given the instance map and green boxes are ground truth bounding boxes.}
\label{fig:detection_example}
\end{figure}

\begin{figure}[t]
\centering
\includegraphics[width=\linewidth]{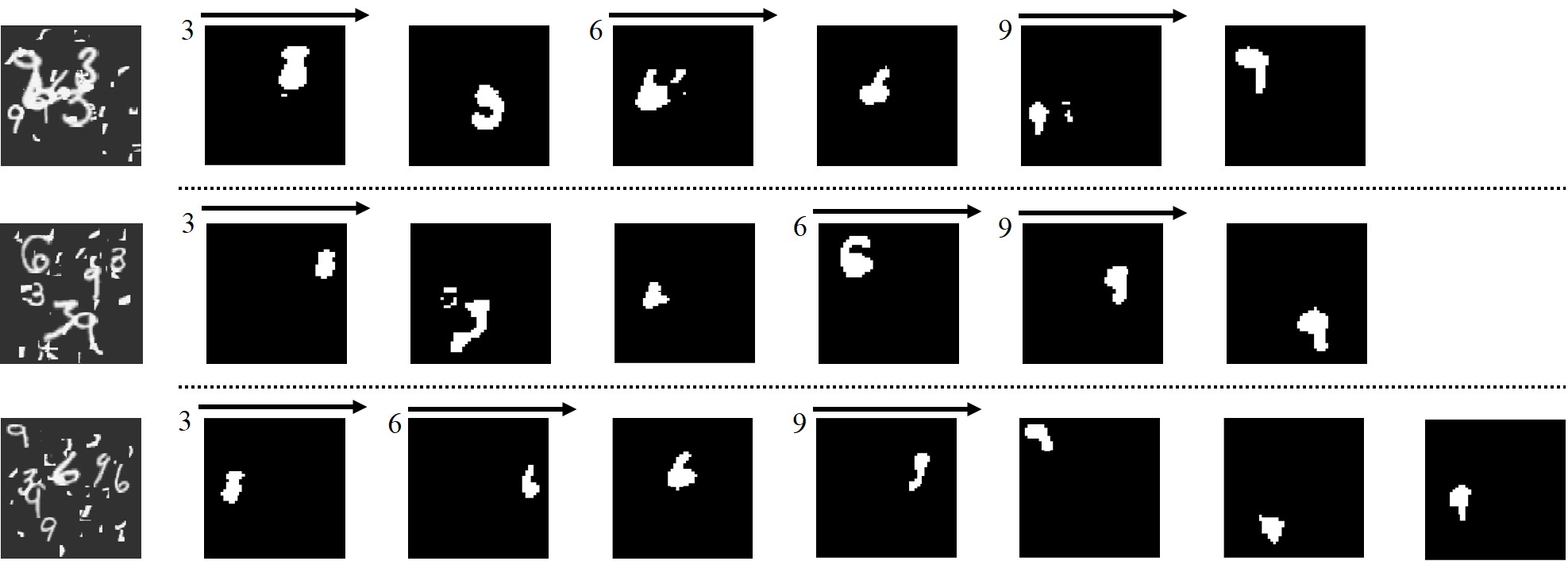}
\caption{Examples of segmentation results: similar to figure \ref{fig:detection_example}. We omitted the map of sum of instance maps and stop signs. We considered response values higher than 0.1 as final segmentation.}
\label{fig:segmentation_example}
\end{figure}

\textbf{Dataset.}
In order to evaluate proposed methods, we synthesized an object detection and instance segmentation dataset based on MNIST dataset. We considered three key aspects in making our dataset. Objects are randomly distributed across an entire image, have various scale, and are allowed to be moderately overlapped with each other. We randomly chose the number of objects from 5 up to 10 per image. The size of each image is 100x100. We selected 3 categories out of 10, which are the digits 3, 6, and 9. Each object in the image was randomly picked from the original MNIST dataset and scaled by a factor of 0.5 up to 2. We allowed overlapping between objects up to 0.2 IoU(Intersection over Union). For realistic scenarios, we added 30 noisy pen strokes per image in the same way as \citet{MnihNIPS2014}. Ground truth labels came for free when we generated the images. Bounding boxes are easily obtained with tight bounds of each object. Segmentation labels are also obtained as a mask whose values are 1 if the corresponding pixel values are greater than zero. Some samples from the dataset are shown in section \ref{fig:detection_example} and \ref{fig:segmentation_example}. We have 30,000 training images and 10,000 testing images.

\textbf{CNN baseline object detection.}
We built and trained a strong baseline method for comparison. We trained a deep CNN for classification with 4 categories, 3, 6, 9, and background. The structure of this CNN is the same as the one we used for category decomposition for fair comparison. During the training, we take random image patches with $\geq 0.5$ IoU overlap with ground truth boxes as positive examples \citep{girshick14CVPR}. Existing region proposal methods didn't work well on our dataset. Thus, we adopted simple yet powerful sliding window approaches. We gave the baseline method a very strong prior. We assumed that the object is square and minimum and maximum size are known. In this way, sliding window with 5 different size square boxes could achieve 99\% recall with step size 5 pixels. We ended up evaluating about 3000 and 1000 boxes for step size 3 and 5 respectively. Finally, we applied non-maximal suppression after evaluating all candidate boxes with threshold 0.3 \citep{girshick14CVPR}.

\textbf{Qualitative study.}
Figure \ref{fig:detection_example} shows some examples of qualitative results for object detection task. In instance maps, we can easily see that our networks put high response values at very precise locations of actual objects. Every time step, our proposed recurrent network tried to produce only one blob. It also handled overlapped instances very well. The overlap between different categories was easily differentiated by our category decomposition network. Overlapping within the same category is handled by a scheme of summing each instance map. Note that category 6 in the first image and category 3 in the third image have overlapped regions and those regions are more highly activated than the other object regions. There are some failures because the network didn't completely succeed in seperating each instance blob. For example, in the 4\textsuperscript{th} instance map of category 9 in the first input image, and the 3\textsuperscript{rd} instance map of category 6 in the third image, they have additional small blobs  that bounding box regression was not able to draw a correct bounding box for.

\begin{figure}[t]
\centering
\begin{subfigure}{.33\textwidth}
  \centering
  \includegraphics[width=\linewidth]{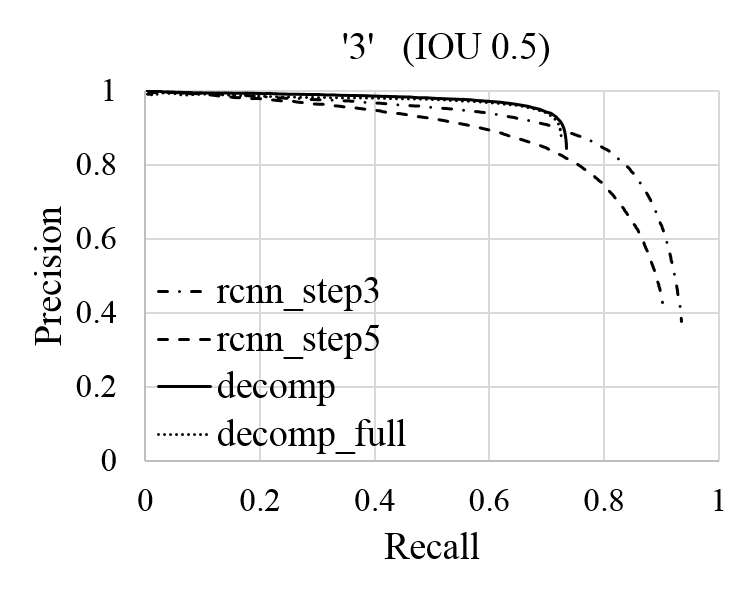}
\end{subfigure}%
\begin{subfigure}{.33\textwidth}
  \centering
  \includegraphics[width=\linewidth]{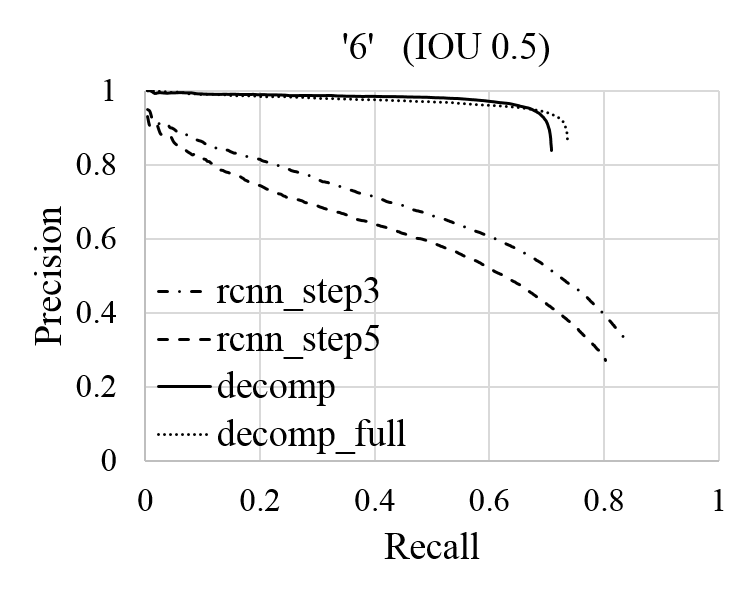}
\end{subfigure}%
\begin{subfigure}{.33\textwidth}
  \centering
  \includegraphics[width=\linewidth]{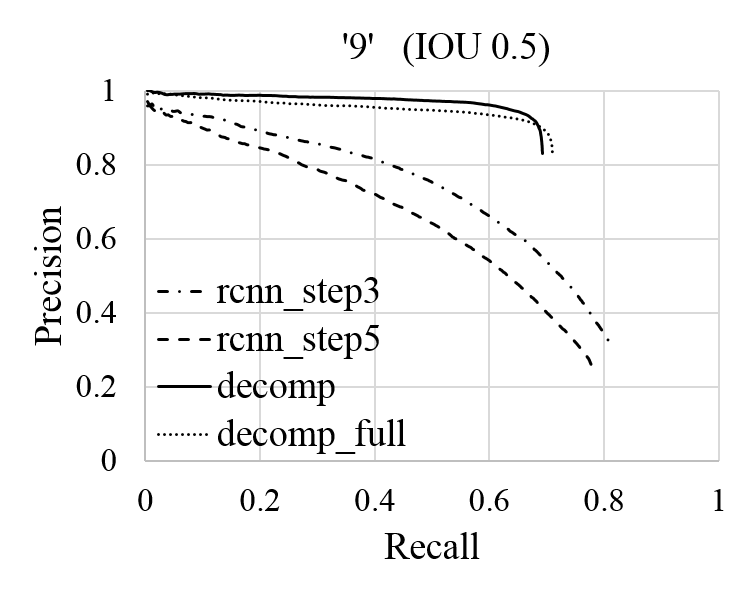}
\end{subfigure}
\caption{Precision-recall (pr) curves of object detection task: Each column shows pr curves for one category with IoU threshold 0.5. \textit{rcnn step3} means baseline detector with step size 3 sliding window and \textit{rcnn step5} means size 5 sliding window. \textit{decomp} is our proposed network with instance maps half the size of the original image. \textit{decomp full} has instance maps same size as the original image.}
\label{fig:pr_curve}
\end{figure}

\begin{figure}[t]
\centering
\begin{subfigure}{.33\textwidth}
  \centering
  \includegraphics[width=\linewidth]{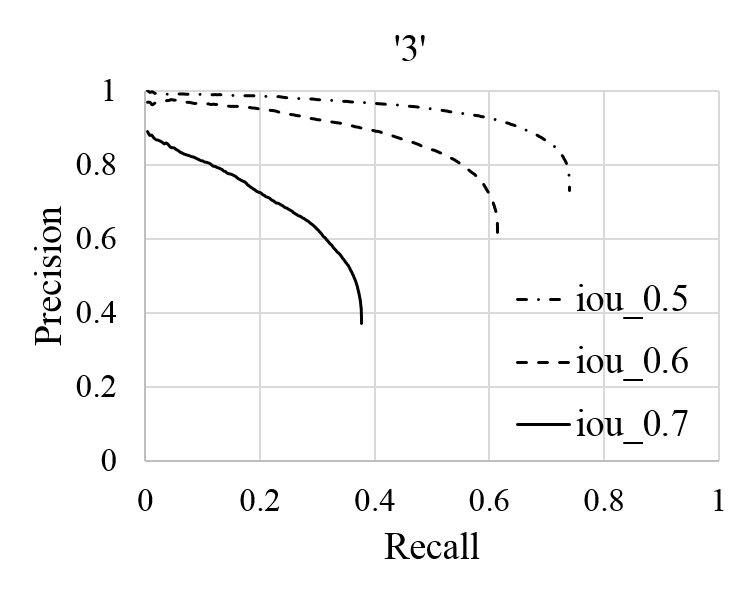}
\end{subfigure}%
\begin{subfigure}{.33\textwidth}
  \centering
  \includegraphics[width=\linewidth]{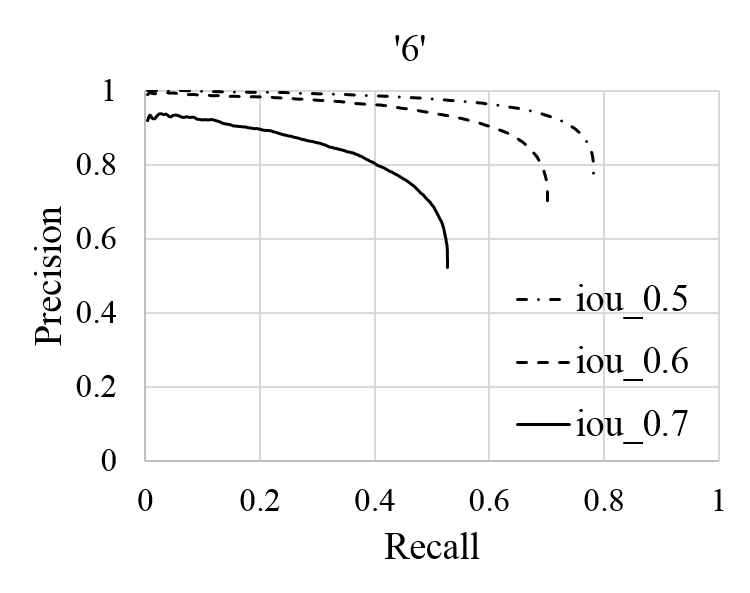}
\end{subfigure}%
\begin{subfigure}{.33\textwidth}
  \centering  
  \includegraphics[width=\linewidth]{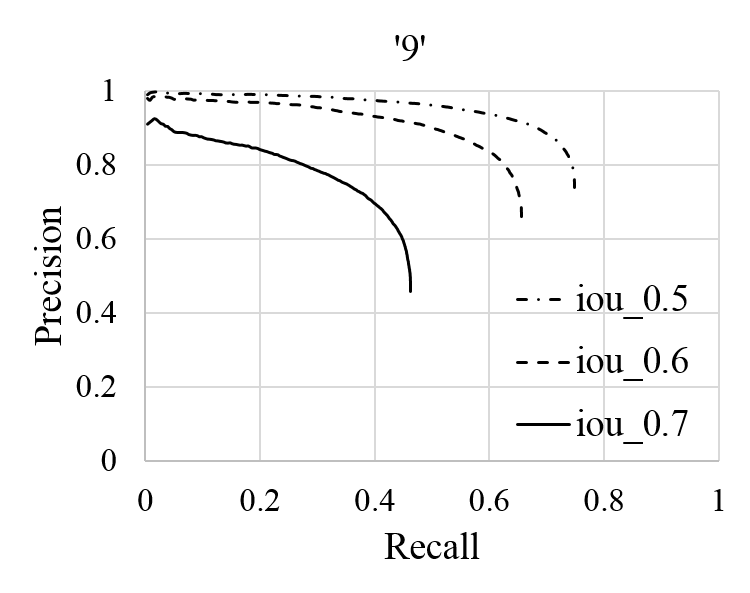}
\end{subfigure}
\caption{Precision-recall (pr) curves of instance segmentation task: similar to figure \ref{fig:pr_curve}. Each graph shows the results of \textit{decomp} for a category evaluated with different IoU thresholds.}
\label{fig:pr_curve_seg}
\end{figure}

\textbf{Quantitative results.}
We computed precision recall curves for comparison with baseline methods(Fig \ref{fig:pr_curve}). For digit 3, our proposed network matched the performance of our strong baseline method with much lower computational cost. We achieved 0.74 recall with very high 0.84 precision. 
It was helpful to produce high resolution instance maps as we noticed from the difference between \textit{decomp} and \textit{decomp full}. For digit 6 and 9, our proposed network outperforms our strong baseline methods. This is mainly because our baseline was confusing these two categories when the image patches were shifted too much from the center of the objects and there was too much noise around the objects.

\textbf{Instance segmentation results.}
Figure \ref{fig:segmentation_example} shows the results of instance segmentation. Not only could our network localize individual object instances separately, but it also segmented out each digits very precisely. However, it often failed to seperate each instance completely if the instances were close. For example, for category 6 in the first input image and category 3 in the second input image, there are unnecessary blobs close to the segmented instances. Figure \ref{fig:pr_curve_seg} shows the precision-recall curve based on IoU metric similar to the detection task. An instance map is correct if its segmentation overlaps with the segmentation of a ground truth instance by more that pre-defined threshold\citep{BharathECCV2014,Liang15}. With same network architecture \textit{decomp}, we could achieve similar performance to the detection task for the category '3' and slightly better performance for the category '6' and '9'. For the segmentation task, we tried individual recurrent networks for each category hoping that we could get better results. However, we couldn't find notable improvement.

\section{Related Work}
\citet{MnihNIPS2014,ba-attention-2015} both proposed visual attention models. They designed the glimpse network that is responsible for explicitly indicating next location of an image to be processed. Then, the classification network processes only a small part of the image associated with the position. We have shown that it is possible to train the network to look at different places at different time steps implicitly without an extra network. The trained network decided where to look by itself and produced an instance map at each time steps. In terms of computational cost, the attention model could save computation since it doesn’t have to process the entire image. However, if there are many objects across the image, it would end up processing the entire area of the image. Furthermore, it might have to process the same regions multiple times if there is overlap between region patches. Our network needs only one evaluation, and repetitive computations in instance decomposition stage are performed by smaller networks.

Very recently, \citet{Liang15} proposed a new instance segmentation method based on feed forward CNN. It is related to our work in terms of the fact that it classified category first, and then divided it into multiple instances. It regresses the number of instance and instance location maps per category. With this information, they applied spectral clustering to separate multiple instances as a post-processing step. In contrast, our network implicitly predicted the number of instances and learned how to cluster response maps, and it is end-to-end trainable. Furthermore, their method is only applicable when there is a segmentation label. However, our method will work as object detection when there is no segmentation label.

\cite{Stewart15} used LSTM to detect multiple people’s faces. It generated multiple instances by looking at only one cell in the spatial grid of the last convolutional feature maps. This would result in running 300(15x20) distinct LSTM for evaluation. They introduced a stitching algorithm to perform non-maximal suppression for dealing with many outputs from many cells. Our method looks at an entire region to produce multiple instances and we don’t require any post-processing steps. And, we can deal with multiple scale objects while they focused only on recognizing very small objects. Finally, their design cannot deal with multiple categories. 

\section{Conclusion and Future work}

Our proposed network has demonstrated it is possible to extract high resolution information for each object instance with a single evaluation. We showed that this information can be used for important computer vision tasks, such as object detection and instance segmentation.

In this paper, we used a regression-based loss function with high resolution masks. Although it has worked quite well, it was not very robust to many learning factors, such as initialization, learning rate, other parameters, and showed slow convergence rate. We believe this is mainly because it has too many degrees of freedom and we could replace it with a pixel-wise classification loss function, which is commonly used in segmentation tasks. In addition, we also strongly believe that spatial regularity of instance maps, such as CRF, would be more effective to achieve better performance and faster learning speed.

We could remove the category decomposition network and attach an instance decomposition network to the CNN directly. However, one of the main motivations for category decomposition is to reduce computational complexity. The recurrent network, which will be running several times according to the number of objects in the image, is currently light weight. All the important jobs can be done by the CNN and category decomposition network. Thus, we can design a relatively small network for the recurrent network and share it across all object categories. Another design choice would be to regress bounding box directly without high resolution instance maps for object detection task. We leave it as future work.

\subsubsection*{Acknowledgments}
We acknowledge support from NSF 1446631, 1452851 and from NVIDIA for GPU hardware. We also would like to thank Hadi Kiapour, Wei Liu, and Phil Ammirato for helpful discussions.

\bibliography{iclr2016_workshop}
\bibliographystyle{iclr2016_workshop}

\section*{Appendix}

\subsection*{A. Object detection on KITTI dataset}
In order to show that our proposed network would work for real and more complicated scenarios, we have tested on KITTI object detection dataset, especially for the car category. KITTI has very high resolution images around 370x1224 and there are many object instances in each image. Some images have more than 15 objects. The size of objects vary from very small objects occupying only 25 pixels in height to large objects occupying more than 300 pixels in height. Furthermore, many objects are highly overlapped each other. Many objects are occluded by more than half of their size.

\begin{figure}
\centering
\includegraphics[width=\linewidth]{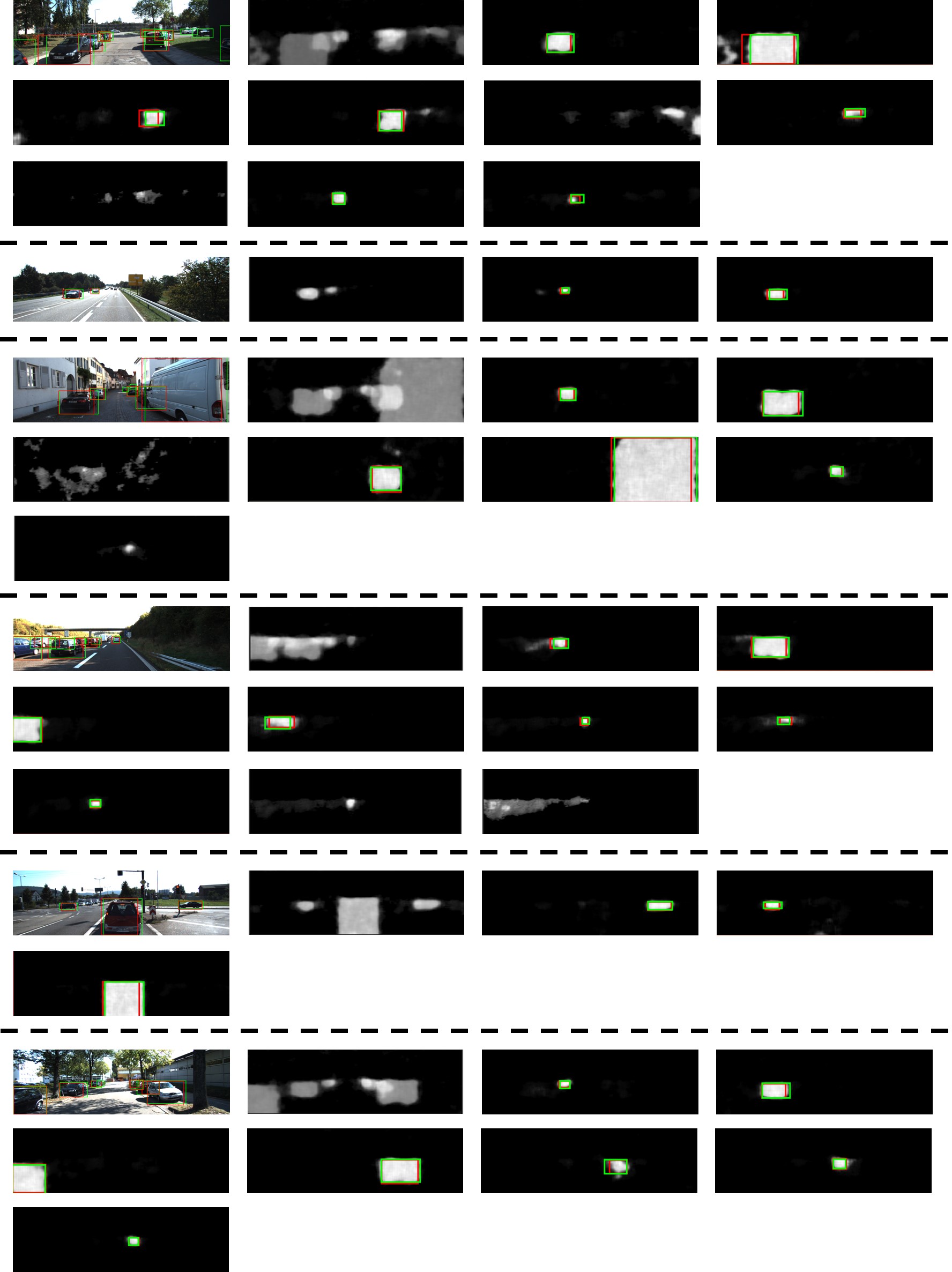}
\caption{Examples of object detection results on the KITTI car dataset: The red boxes are estimated bounding boxes and the green boxes are ground truth. The first column is input image and the second column is the sum of all instance maps. The instance map without any boxes indicates false positive.}
\label{fig:kitti_example}
\end{figure}

Figure \ref{fig:kitti_example} shows good examples of car detection. One interesting thing we found out is there is specific ordering during the instance decomposition process. During the training process, we didn't specify any ordering of instances. However, the recurrent network first looks at the center of the image. Then it goes to the left side of the image and then to the right. We observed this phenomenon across many images. This is mainly because many images have similar patterns(e.g. large cars on both sides of the image, and smaller cars in the middle of the image) and the network found optimal ordering for those patterns. This was not the case for the synthesized digit dataset because it was randomly generated.

We admit that our current simple recurrent network does not achieve very good results for large number of objects in an image. In KITTI dataset, some images have more than 10-20 cars in an image. In this case our network performed poorly. We plan to apply LSTM style network for dealing with long term dependency and better performance. In addition, we also suffered from recognizing small objects. We also plan to apply weighted loss function for activation maps in order to consider the size of the objects.

\subsection*{B. Implementation details}
Deconvolutional layer is defined as flipped version of convolutional layers and introduced in \citet{long_shelhamer_fcn}. \textit{conv} and \textit{deconv} has 5 parameters, which are the number of input and output channels, filter size, and stride. The number hidden units in fully connected recurrent layer\textit{rnn-fc} was set to 2048. Batch normalization and rectifier linear unit were used after every layers. The stride 2 was used for down and upsampling. The network architecture used in this paper for synthesized dataset is following. The category decomposition network is \textit{conv(1,32,5,2)}-\textit{conv(32,32,5,2)}-\textit{conv(32,64,3,2)}-\textit{conv(64,64,3,1)}-\textit{conv(64,64,3,2)}-\textit{conv(64,64,3,1)}-\textit{deconv(64,64,3,2)}-\textit{deconv(64,64,3,2)}-\textit{deconv(64,64,3,2)}-\textit{deconv(64,3,1,1)}. The instance decomposition network is \textit{conv(1,32,5,2)}-\textit{conv(32,32,3,2)}-\textit{conv(32,32,3,2)}-\textit{rnn-fc}-\textit{rnn-fc}-\textit{deconv(32,32,3,1)}-\textit{deconv(32,32,3,1)}-\textit{deconv(32,32,5,1)}-\textit{deconv(32,1,1,1)}.

We used standard stochastic gradient descent method and performed two stage training procedure. First we trained category decomposition network with the loss function described in section \ref{sec:category_decomposition}). Once we trained, we fixed it when we train the second part of network with loss function described in section \ref{sec:instance_decomposition}. Finally, we combined both together and end-to-end trained the whole network. We could train the whole network from the scratch. However, it was very sensitive to the setting of hyperparameters. The hyperparameter $\lambda$ and $\gamma$ was important. If it is too small, it decreased learning speed. However, if it is too large the network gave trivial solution with producing all zeros. So, we first used small value of $\lambda = \gamma = 0.3$, after 1 epochs, we increased $\lambda = \gamma =1.0$. For $\eta$ we set to 1 in every experiments. 

We trained category decomposition network first. And we trained instance decomposition network without updating category network. And then, we fine-tuned all networks together. For KITTI dataset, we used pretrained VGG16 network\citep{verydeep}. We attach deconvolutional layers to it. Similarly, we trained each network seperately, and fine-tuned the network all the way down to the first convolutional layer of VGG16 network. All implementations were based on torch deep learning library\citep{torch}.

\begin{figure}
\centering
\includegraphics[width=\linewidth]{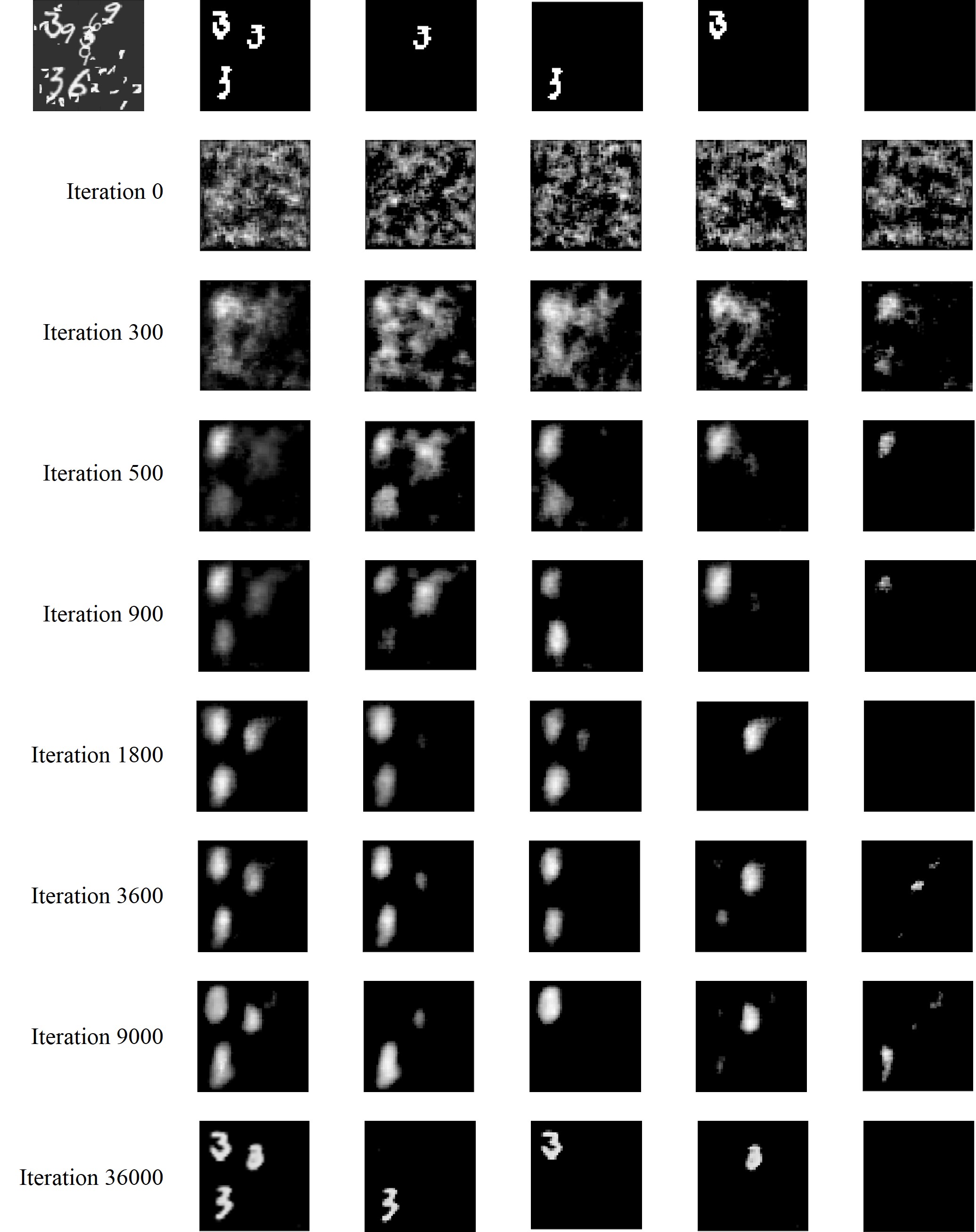}
\caption{Training evolution over iterations: The first row shows input image and corresponding ground truth mask of digit '3' for instance segmentation. From the second row to the last row, the second column shows the sum of instance maps that our network produced. From the third to the last column shows each instance map at each time step.}
\label{fig:training_evolution}
\end{figure}

\begin{figure}
\centering
\includegraphics[width=\linewidth]{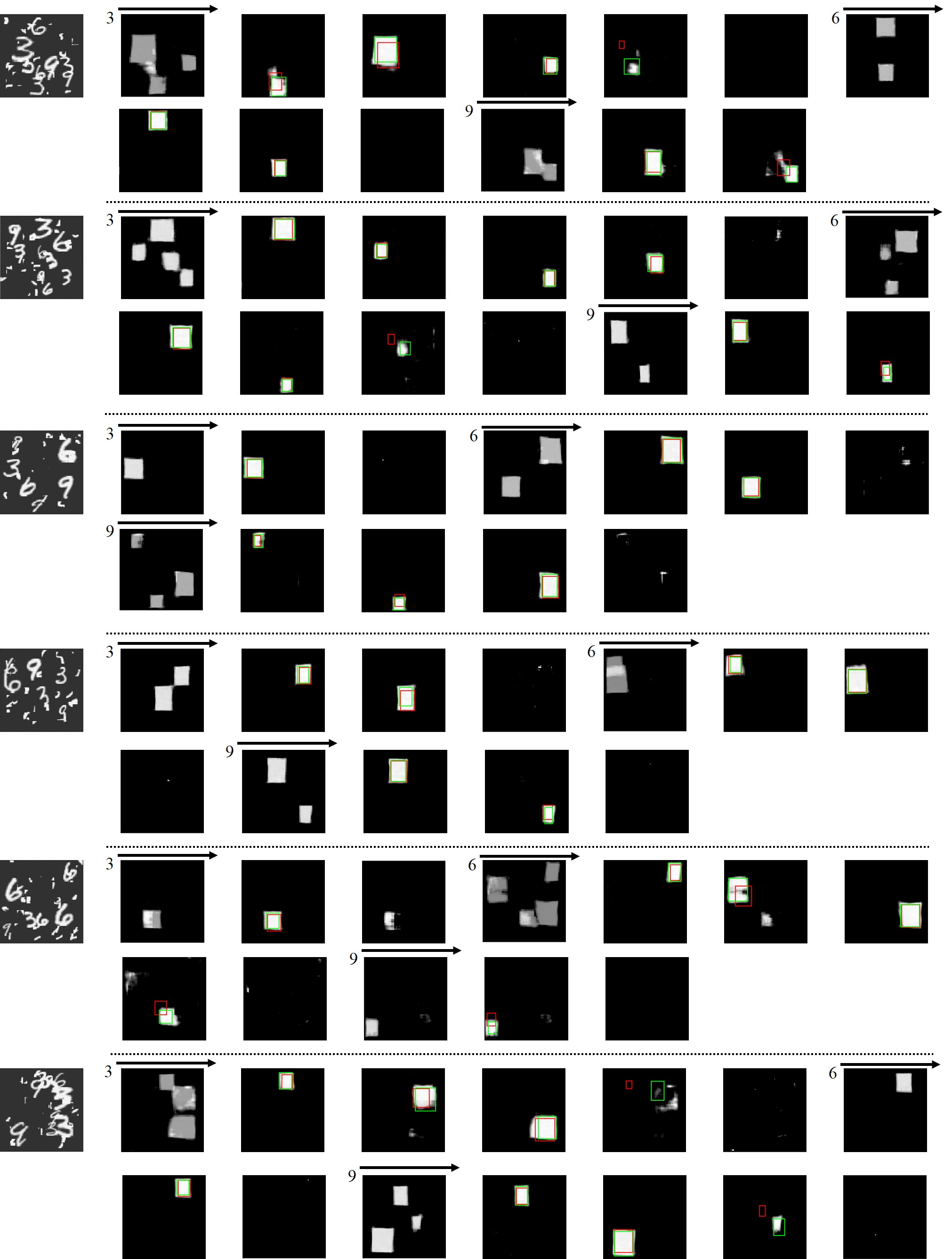}
\caption{More examples of object detection results}
\label{fig:detection_example_appendix}
\end{figure}

\begin{figure}
\centering
\includegraphics[width=\linewidth]{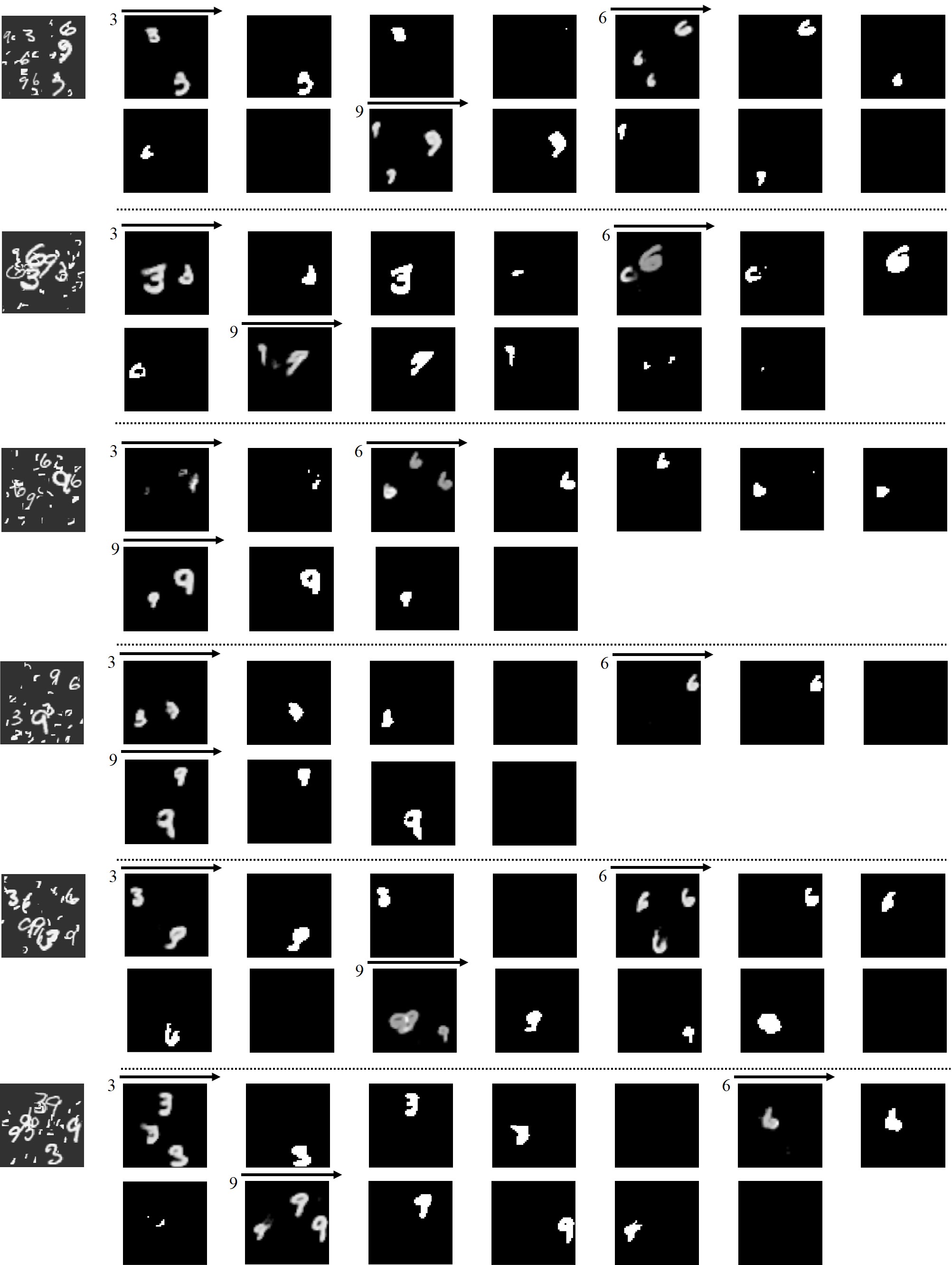}
\caption{More examples of instance segmentation results}
\label{fig:segmentation_example_appendix}
\end{figure}

\end{document}